\documentclass{article}

\usepackage{ijcai17}

\usepackage{times}

\usepackage{algorithm}
\usepackage{algorithmicx}
\usepackage{algpseudocode}
\usepackage{mathtools}
\usepackage{url}
\usepackage{graphicx}

\title{Improving classification accuracy of feedforward neural networks for spiking neuromorphic chips}

\author{}

\author{Antonio Jimeno Yepes, Jianbin Tang, Benjamin Scott Mashford\\
        antonio.jimeno@au1.ibm.com, jbtang@au1.ibm.com, bmashfor@au1.ibm.com \\
        IBM Research Australia, Carlton, 3053, VIC, Australia}

\begin{document}

\maketitle

\begin{abstract}

Deep Neural Networks (DNN) achieve human level performance in many image analytics tasks but DNNs are mostly deployed to GPU platforms that consume a considerable amount of power.
New hardware platforms using lower precision arithmetic achieve drastic reductions in power consumption.
More recently, brain-inspired spiking neuromorphic chips have achieved even lower power consumption, on the order of milliwatts, while still offering real-time processing.

However, for deploying DNNs to energy efficient neuromorphic chips the incompatibility between continuous neurons and synaptic weights of traditional DNNs, discrete spiking neurons and synapses of neuromorphic chips need to be overcome.
Previous work has achieved this by training a network to learn continuous probabilities, before it is deployed to a neuromorphic architecture, such as IBM TrueNorth Neurosynaptic System, by random sampling these probabilities.

The main contribution of this paper is a new learning algorithm that learns a TrueNorth configuration ready for deployment. We achieve this by training directly a binary hardware crossbar that accommodates the TrueNorth axon configuration constrains and we propose a different neuron model.

Results of our approach trained on electroencephalogram (EEG) data show a significant improvement with previous work (76\% vs 86\% accuracy) while maintaining state of the art performance on the MNIST handwritten data set.

\end{abstract}

\section{Introduction}

Neural networks achieve human level performance in many image analytics tasks but neural networks are mostly deployed to GPU platforms that consume a considerable amount of power.
New hardware platforms using lower numerical precision achieve drastic reductions in power consumption.
Brain-inspired spiking neuromorphic chips such as the IBM TrueNorth Neurosynaptic System~\cite{merolla2014million} have achieved power consumption on the order of milliwatts while still offering high accuracy and real-time processing.
Recent work~\cite{esser2015backpropagation,esser2016convolutional} has used these ideas to implement constrained neural networks to enable high performance image analytics on TrueNorth~\cite{merolla2014million}.
With recent advances in hardware platforms already available, there is an urge to provide software that fully exploits the potential of this new type of hardware.

Recent approaches in machine learning have explored training models constrained to binary weights~\cite{courbariaux2015binaryconnect} or low precision arithmetic~\cite{courbariaux2014low,cheng2015training,hwang2014fixed} and spiking deep belief neural networks~\cite{stromatias2015scalable}, achieving state of the art performance in image analytics tasks.
Lower power consumption is achieved by using only additions (avoiding expensive multiplication modules), which can benefit from implementing these models in configurable architectures such as FPGA or brain-inspired systems.


Recent methods for training adaptations of fully connected style layers to the TrueNorth core-to-core connectivity~\cite{esser2015backpropagation} applied to EEG data have shown that results using neuromorphic platforms exhibit performance levels that are below current work using unconstrained hardware~\cite{nurse2016acm}.
In this work, we revise the training of deep neural networks for low energy neuromorphic architectures and demonstrate high performance levels comparable to non-constrained neural networks on MNIST and EEG data sets.

The selection of datasets in this study is guided by emerging areas of application for neuromorphic computing platforms. It has been proposed that the development of high-performance classifiers that run within a constrained power consumption limit will enable a new generation of mobile sensing applications~\cite{lane2015early}. Early examples of such applications include healthcare-related wearables that involve machine vision and audio-signal processing~\cite{poggi2016wearable}, as well as real-time analysis of  imagery for visual monitoring and surveillance ~\cite{antoniou2016general}.  




This paper is organized as follows.
First feedforward neural networks and their training are introduced, followed by an overview of the TrueNorth chip architecture.
Then a revised constrained model of the feedforward neural network training and deployment for TrueNorth are introduced.
Finally results on EEG data and the MNIST handwritten data set are presented with discussion and future work.

\section{Feedforward Neural networks}

In feedforward neural networks, inputs $x_i$ are integrated in units $I_j$ using a set of weights $w_{ij}$ and a bias term $b_j$ (which can be shared among units) as shown in equation~\ref{eq:integration}.
During training, the weights $w_{ij}$ and bias $b_j$ are learnt.

\begin{equation}
I_j = \sum_i x_i w_{ij} + b_j
\label{eq:integration}
\end{equation}

\begin{figure}[!ht]
  \centering
  \includegraphics[width=0.85\columnwidth]{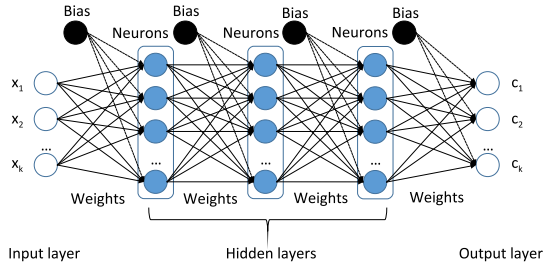}
  \caption{Neural network}
  \label{fig:neural-network}
\end{figure}

Neurons output are a combination of the integration function $I_j$ followed by a non-linearity, which can be a sigmoid function.
The neurons can be grouped into layers and these layers can be stacked as shown in figure~\ref{fig:neural-network}.

Prediction is done using the output layer outcome.
In classification, typically each neuron is associated to one prediction category.
Information is integrated in the output layer neurons, which are then be turned into probabilities using, for instance, softmax (eq.~\ref{eq:softmax}).
$p_l$ is the prediction probability for category $c_l$ from all the possible $K$ categories.

\begin{equation}
p_l = \frac{e^{c_l}}{\sum^{K}_{k=1} e^{c_k}}
\label{eq:softmax}
\end{equation}

During training, weight and bias values are typically learnt using backpropagation~\cite{duda2012pattern}, which minimizes a loss function that represents classification error of the network using gradient descent.
An example of such a loss function is the logarithmic loss, which is used in this work (eq.~\ref{eq:loss}), even though other functions could be explored.

\begin{equation}
E=-\sum_k \lbrack y_k log(p_k) + (1-y_k) log (1-p_k)\rbrack
\label{eq:loss}
\end{equation}

The next sections describe the neuromorphic hardware and an explanation about how the neural networks can be constrained, learnt and deployed. 

\subsection{Deployment hardware}

IBM TrueNorth~\cite{merolla2014million}, a low power and highly parallelized brain-inspired chip is used here as an example deployment system, though our method could be generalized to other neuromorphic hardware~\cite{benjamin2014neurogrid,painkras2013spinnaker,pfeil2012six}.
In its current implementation it is composed of 4,096 neurosynaptic cores.

\begin{figure}[!ht]
  \centering
  \includegraphics[width=0.7\columnwidth]{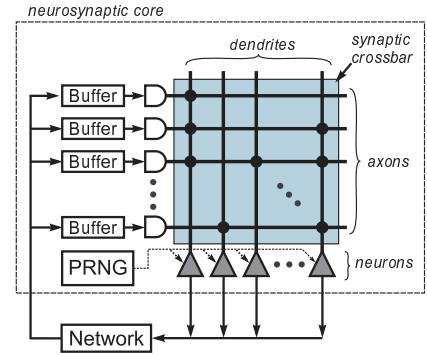}
  \caption{TrueNorth neurosynaptic-core}
  \label{fig:truenorth_core}
\end{figure}

Figure~\ref{fig:truenorth_core} shows the layout of one of these cores.
Each core has highly configurable 256 input axons and 256 output neurons. 
Input spikes enter the core through axons and the information is passed via the  configurable binary synaptic crossbar to the leaky integrate-and-fire neurons, which act as computation units. The neuron assigns weights to the input spikes, integrates and updates its membrane potential, compares it to a threshold and may spike and reset, based on the selected threshold, reset, and stochastic modes~\cite{cassidy2013neuronmodel}.

Programming for TrueNorth means writing core configurations and connections, which in a way could be seen as building a neural network.
A programming language has been developed to abstract the chip configuration~\cite{amir2013cognitive}.
In our case, the learned neural networks need to be translated into a valid TrueNorth configuration.

In previous work~\cite{esser2015backpropagation}, a constrained network is trained, which learns the crossbar connection probability and biases.
To deploy the model into TrueNorth these probabilities are sampled into discrete synaptic crossbar connections.

\section{Constrained training model}

Neuromorphic chips such as TrueNorth provide a reduction of power consumption, which is achieved by a constrained hardware architecture.
There are several constraints compared to more traditional hardware platforms that are explored in this section to define the training of neuromorphic chips.
These constraints are: use of spiking neurons, weights constrained to a limited set of integer values and reduced connectivity.
In the following sections, we describe these constraints in more detail and provide a way to do learning based on backpropagation.
We also describe how to use output layer predictions integrated into backpropagation.

\subsection{Spiking neurons}

Spiking neurons have two possible outputs (0, no output, or 1, spike) as shown in the revised neuron output shown in eq.~\ref{eq:neuron-output}.

\begin{equation}
n_j =
  \begin{cases}
    1, & \mbox{if } I_j > 0 \\
    0, & \mbox{otherwise}
  \end{cases}
\label{eq:neuron-output}
\end{equation}

Spiking neurons behavior could be seen as a Heaviside step function, which does not have a gradient for backpropagation since it is not differentiable.
This step function can have many different smooth approximations using, for instance, the logistic function that is used to represent the neuron output $\hat{n}_j$ as shown in eq.~\ref{eq:logistic}.
The parameter $k$ is the steepness of the function, which in our work has been set empirically to $1/2$.
Other values of $k$ could be explored or, may be, approximate its value to a logistic distribution using large data sets.

\begin{equation}
\hat{n}_j \approx \frac{1}{1 + e^{-2kI_{j}}}
\label{eq:logistic}
\end{equation}

The following gradient formulation can then be used in backpropagation.

\begin{equation}
\frac{\partial \hat{n}_j}{\partial I_j} = \frac{1}{1 + e^{-2kI_{j}}} * (1 - \frac{1}{1 + e^{-2kI_{j}}}) 
\end{equation}

In this approach, each neuron is stateless, so its status is reset after each time cycle, or tick. This is implemented in TrueNorth by setting parameters as specified and each neuron’s leak equal to the rounded bias term, with its threshold to 0, its membrane potential floor to 0~\cite{esser2015backpropagation}.

\subsection{Constrained weights}

In TrueNorth, effective weights $w_{ij}$  are divided in two components: the axon weight $s_{ij}$ and the crossbar connection $c_{ij}$, so $w_{ij} = c_{ij} s_{ij}$.
Taking this into account, the integration function is defined as:

\begin{equation}
\hat{I}_j = \sum_i x_i c_{ij} s_{ij} + b_j
\label{eq:truenorth-aggregation}
\end{equation}

Instead of learning the weights $w_{ij}$, the axon weights are proportionally distributed as explained below the binary crossbar connection $c_{ij}$ is learnt during training.

The hardware accepts two possible values for the crossbar connections (0,1).
The effective weights are dependent on the configuration of the axon types and crossbar connections.
In this setup, the total number of effective weights that can be used is a maximum of five, i.e. four axon type weights and no connection (which means zero weight).
The distribution of weights in $s_{ij}$ depends on the axon type configuration in a TrueNorth core (we follow~\cite{esser2015backpropagation}).

The backpropagation gradients are specified as below and a similar formulation is used for bias values, which are round to the closer integer value during deployment.
In the next section, we explain how the binary crossbar connections are learnt.

\begin{equation}
\frac{\partial \hat{n}_j}{\partial c_{ij}} = \frac{\partial \hat{n}_j}{\partial \hat{I}_j} \frac{\partial \hat{I}_j}{\partial c_{ij}} ; \frac{\partial \hat{I}_j}{\partial c_{ij}} = x_i s_{ij}
\end{equation}

\subsection{Binary crossbar connection learning}

In the previous section, we have shown that binary crossbar connections need to be learnt.
We propose adapting previous work~\cite{esser2016convolutional,courbariaux2015binaryconnect} to learn binary crossbar connections that are used to obtain the effective network weights as shown in algorithm~\ref{alg:training}.

This is in contrast to previous work in which connection probabilities are learnt~\cite{esser2015backpropagation} that need to be sampled prior to hardware deployment.
As well, in contrast to previous work~\cite{esser2016convolutional,courbariaux2015binaryconnect}, the proposed approach allows learning a larger number of weights that are defined by the hardware configuration.

\begin{figure}[!ht]
  \centering
  \includegraphics[width=1.0\columnwidth]{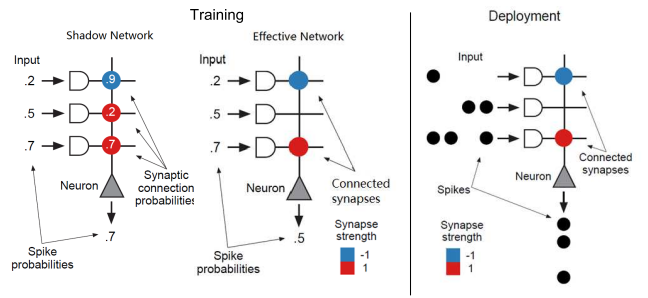}
  \caption{Crossbar connections during training and deployment}
  \label{fig:crossbar-learning}
\end{figure}

During training, in the forward propagation step the high precision $c^{sh}$ matrix from the shadow network is used to calculate the $c$ binary connection values (eq.~\ref{eq:cbin}), which are used to calculate the effective network weights using ($w_{ij} = c_{ij} s_{ij}$).
Effective network weights are used to calculate the performance of the trained model (c.f. figure~\ref{fig:crossbar-learning}).

\begin{equation}
c_{ij} =
  \begin{cases}
    1, & \mbox{if } c^{sh}_{ij} > 0.5 \\
    0, & \mbox{otherwise}
  \end{cases}
	\label{eq:cbin}
\end{equation}

During backpropagation, gradients are estimated using $c$ connection values that are used to update the $c^{sh}$ high precision connection probabilities.


%

\begin{algorithm}
\caption{Forward and backward propagation}
\label{alg:training}
\begin{algorithmic}
  \State{//Forward propagation}
	\For {$l=1:layers$}
	  \For {$c=1:layer(l).cores$}
	    \State Calculate core binary crossbar connections $c_{ij}$ (c.f. eq.~\ref{eq:cbin}) and effective weights $w_{ij}$
	    \State Calculate core neuron outputs $\hat{I}_j$ using $c_{ij}$
		\EndFor
	\EndFor
	\State Estimate log loss in the last layer
	\State{//Backpropagation}
  \For {$l=1:layers$}
	  \For {$c=1:layer(l).cores$}
      \State Calculate gradients using $c_{ij}$
	    \State Update $c^{sh}_{ij}$ using estimated gradients
		\EndFor
	\EndFor
\end{algorithmic}
\end{algorithm}

\subsection{Neural network connections}

The described method allows defining fully connected layer networks, but TrueNorth cores have 256 inputs (axons) and 256 outputs (neurons), which constrains the size of fully connected layers.
When designing a network, the first layer, which processes the input data, is defined by blocks that tile to cover the input data.
Block size and stride are defined by the number of pixels in the image classification case.
Since each of these blocks maps to a TrueNorth core, the number of input axons required to map the input data per block is calculated as $({blockSize}^2 * numberChannels)$ and the maximum value is 256, which is the number of axons per core.
In the upper layers, block size and stride are defined by the number of cores (instead of image size as in the input layer).
Output neurons of the previous layer are mapped to the input axons of the cores in the upper layer.

\subsection{Output layer}

The last layer of the of the network is used for prediction. 
Spiking neurons in the last layer can be assigned randomly to one category~\cite{esser2015backpropagation}.
The category with the highest number of neurons firing will be the predicted one.

These counts need to be used in backpropagation.
In our work, we have used the number of spiking neurons per category $cs_{k}$ divided by the number of neurons per category $cn_{k}$ and used~\textit{softmax} as shown in equation~\ref{eq:softmax-tn} to estimate the per category probability that is then used with the logarithmic loss (eq.~\ref{eq:loss}).

\begin{equation}
\hat{p}_l = \frac{e^{\frac{cs_{l}}{cn_{l}}}}{\sum^{K}_{k=1} e^{\frac{cs_{k}}{cn_{k}}}}
\label{eq:softmax-tn}
\end{equation}

We have described neural networks, the constraints added by neuromorphic chips and how to adapt the training of neural networks to these constraints.
In the following sections, we present the results obtained using the MNIST data set and EEG data.

\section{Results}

We have used two data sets in our experiments.
An EEG data set, described in more detail below, and the MNIST handwritten data set. 

EEG data\footnote{https://github.com/EwanNurse/A-Generalizable-BCI-using-Machine-Learning-for-Feature-Discovery} was recorded from a participants wearing electrodes placed on the scalp while they performed a self-paced hand squeeze task. Full experimental details are described elsewhere~\cite{nurse2015generalizable}. Briefly, the participant wore surface electromyography (EMG) electrodes attached to the forearms that were used to detect the onset of each hand squeeze. Data was sampled at 1 kHz in  four minute sessions, with a total recording time of approximately 30 minutes.  Detection of hand squeeze onset was achieved by extracting the root mean squared (RMS) power from the EMG signal, and the corresponding EEG data was labeled with either left-hand squeeze or right-hand squeeze (2 classes).
Each EEG instance has been translated to 46x46 images with one 8-bit channel.
In this study, EEG signals from participant D are used, which allows comparing the results to previous work.
This data set contains 480/468 examples per class in the training set and 66/95 examples per class in the testing set.
The data set consisted of a randomly allocated unseen test set (20\% of the data) and training set (80\% of the data), allowing a direct comparison with previous results~\cite{nurse2015generalizable}.

MNIST data set~\cite{lecun1998gradient} contains handwritten digits from 0 to 9 with 60,000 examples for training and 10,000 samples for testing.
Images are 28x28 pixels, with one 8-bit gray scale channel.
Gray scale values have been normalized dividing them by 255.

During training EEG and MNIST data are represented using floating point values.
During deployment, the data values a single input axon over n ticks, which means that floating point values need to be represented as a set of spikes.
Rate code~\cite{Jimeno2016weighted} has been used to encode the data.
In rate code, the number of spikes in the time window represents the encoded value, k.
Spikes are evenly distributed within the time window.

Experiments were run using a MATLAB implementation of the proposed method, this includes training and evaluation of the trained models in a simulated MATLAB environment.
We have used two configurations for the axon type weights in the neuron configuration s=[-1,1], used in previous work, and s=[-2,-1,1,2].

Table~\ref{tab:networks} lists the networks used in the experiments.
Network configurations are derived from previous work~\cite{esser2015backpropagation}.
There is a small network with 3 layers that requires 9 TrueNorth cores and a larger network with 4 layers that requires 30 TrueNorth cores.

\begin{table}[ht]
\begin{centering}
\begin{tabular}{ccc}
\hline
Name         & Layer & Definition                 \\
\hline
Small network& 1     & Block size 16 Stride 12    \\
             & 2     & Block size 1  Stride 1   \\
             & 3     & Block size 2  Stride 1   \\
\hline						
Large network& 1     & Block size 16 Stride 4 \\
             & 2     & Block size 2  Stride 1 \\
			       & 3     & Block size 2  Stride 1 \\
			       & 4     & Block size 2  Stride 1 \\
\hline
\end{tabular}
\par
\end{centering}
\caption{Networks used in the experiments.
         MNIST small has 3 layers and requires 9 TrueNorth cores.
         MNIST large has 4 layers and requires 30 TrueNorth cores.}
\label{tab:networks}
\end{table}

Performance of the deployed trained models has been measured using classification accuracy, which is calculated by dividing the number of correctly predicted instances by all instances in the test set.
McNemar's test has been used to look if the outcome of the systems is statistically significant~\cite{haralick1993performance}.
During test, data is converted into spikes using a rate code scheme.
Since rate code is used and there is no need to sample from the trained network crossbar probabilities, results are stable (e.g. repetitions of the experiments provide the same result) compared to previous methods.

We have found that the initial optimal learning rate $lr$ was provided by equation~\ref{eq:learning-rate}, which takes into account the number of neurons assigned to each category and the batch size.

\begin{equation}
lr = (neurons\_per\_category * batch\_size)^{-1}
\label{eq:learning-rate}
\end{equation}

\subsection{EEG results}

For EEG experiments, the initial learning rate $lr$ was multiplied by 0.1.
We have updated the learning rate every 100 iterations using a rate of 0.1.
Batch size is 25 and we have run a total of 500 epochs.

Ranges of values vary a lot for EEG data across features.
We have normalized each individual feature $x_i$ but instead of using values between -1 and 1 as it is typically done with unconstrained neural networks~\cite{ioffe2015batch}, the values have been normalized to be between 0 and 1 as shown in the formula below.
A small $\epsilon$ value has been inserted to avoid zero values.

\begin{equation}
\hat{x_i}= \frac{\frac{x_i - E[x^{train}_i]}{\sqrt{var(x^{train}_i) + \epsilon}}}{2}+0.5
\end{equation}

Results are shown in Table~\ref{tab:results-eeg} training the network with two different axon type weights and the combination of both trained models.
Previously reported results using TrueNorth showed a 76\% accuracy for participant D on this 2-class task~\cite{nurse2016acm}.
With the proposed method we obtain over 86\%, which is the same accuracy as previously reported for the same data set using an unconstrained neural network~\cite{nurse2015generalizable}.
Better results are obtained when a more restricted set of axon type weights ($s=[-1,1]$) is used.
Normalization of the set at the feature level improves the performance of the trained models, which indicates that feature preprocessing can benefit predictive performance.
Both results are statistically significant ($p < 0.005$).
Using more than 32 ticks does not show a statistically significant improvement.

\begin{table}[ht]
\begin{small}
\begin{centering}
\begin{tabular}{c|cc|cc}
\hline
     &\multicolumn{2}{c|}{s=[-2,-1,1,2]}&\multicolumn{2}{c}{s=[-1,1]}\\
\hline
T    &Original       &Normalized     &Original       &Normalized     \\
\hline
1    &61.43$\pm$3.72 &76.58$\pm$2.17 & 62.36$\pm$5.27&79.19$\pm$2.45 \\
2    &64.66$\pm$4.82 &78.01$\pm$1.58 & 66.96$\pm$4.22&81.80$\pm$2.43 \\
4    &68.94$\pm$2.40 &80.93$\pm$1.78 & 70.06$\pm$3.98&84.22$\pm$1.50 \\
8    &71.12$\pm$2.22 &81.30$\pm$2.46 & 72.36$\pm$2.92&84.66$\pm$1.58 \\
16   &73.73$\pm$3.21 &82.55$\pm$1.48 & 75.96$\pm$2.17&84.07$\pm$1.40 \\
32   &76.27$\pm$3.14 &83.17$\pm$0.99 & 80.00$\pm$2.34&85.90$\pm$0.66 \\
64   &76.52$\pm$2.23 &83.60$\pm$0.52 & 80.62$\pm$1.89&{\bf 86.09$\pm$0.73} \\
\hline
\end{tabular}
\par
\end{centering}
\end{small}
\caption{EEG accuracy results using the large network and two axon weights $s$ configuration and with (Normalized) and without normalization (Original).
T stands for the number of ticks used to encode test instances with rate code.}
\label{tab:results-eeg}
\end{table}

\subsection{MNIST results}

Training has been carried out using batches of 100 images for each iteration and with and without data augmentation.
To perform data augmentation, random changes to the images have been done in each epoch.
We used a maximum rotation of 7.5 degrees, a maximum shift of 2.5 and a maximum rescale of 7.5\%.
Augmentation changes are randomly performed.
For experiments without augmentation, the initial learning rate $lr$ is multiplied by 0.1 and 500 epochs were used.
For augmentation experiments, the initial learning rate $lr$ is multiplied by 0.1, which is further multiplied by 0.1 after 2,000 epochs.
3,000 epochs were used for training.
More iterations were required for the training set using augmentation to converge compared to the number of iterations used without augmentation.

Tables~\ref{tab:results-mnist-small} and~\ref{tab:results-mnist-large} show the results for several configurations.
The large network shows better performance compared to the smaller one.
The configuration with axon types ($s=[-1,1]$) provide better performance.
Augmentation increases the performance, which is similar to previous published work~\cite{esser2015backpropagation}.
Improvements in performance by these three configurations are statistically significant ($p < 0.005$).
As with the EEG results, using more than 32 ticks does not show a statistically significant improvement.

\begin{table}[ht]
\begin{small}
\begin{centering}
\begin{tabular}{c|cc|cc}
\hline
     &\multicolumn{2}{c|}{s=[-2,-1,1,2]}&\multicolumn{2}{c}{s=[-1,1]}\\
\hline
T    &No aug        &Aug           &No aug        &Aug           \\
\hline
1    &95.05$\pm$0.14&95.99$\pm$0.13&93.73$\pm$0.21&95.20$\pm$0.16\\
2    &96.27$\pm$0.09&97.46$\pm$0.13&96.07$\pm$0.20&97.17$\pm$0.11\\
4    &97.00$\pm$0.09&98.20$\pm$0.08&97.28$\pm$0.11&98.22$\pm$0.11\\
8    &97.30$\pm$0.09&98.44$\pm$0.03&97.75$\pm$0.06&98.58$\pm$0.09\\
16   &97.44$\pm$0.06&98.56$\pm$0.07&97.95$\pm$0.08&98.76$\pm$0.05\\
32   &97.56$\pm$0.07&98.64$\pm$0.05&98.03$\pm$0.05&98.84$\pm$0.02\\
64   &97.65$\pm$0.04&98.66$\pm$0.03&98.08$\pm$0.04&{\bf 98.86$\pm$0.03}\\
\hline
\end{tabular}
\par
\end{centering}
\end{small}
\caption{MNIST small network results using data augmentation (Aug) and not using it (No aug).
T stands for the number of ticks used to encode test instances with rate code.}
\label{tab:results-mnist-small}
\end{table}

\begin{table}[ht]
\begin{small}
\begin{centering}
\begin{tabular}{c|cc|cc}
\hline
     &\multicolumn{2}{c|}{s=[-2,-1,1,2]}&\multicolumn{2}{c}{s=[-1,1]}\\
\hline
T    &No aug        &Aug           &No aug        &Aug           \\
\hline
1    &95.45$\pm$0.13&96.97$\pm$0.13&95.36$\pm$0.17&97.26$\pm$0.13\\
2    &96.50$\pm$0.11&98.07$\pm$0.11&96.45$\pm$0.12&98.32$\pm$0.10\\
4    &97.16$\pm$0.06&98.60$\pm$0.07&97.38$\pm$0.10&98.88$\pm$0.07\\
8    &97.53$\pm$0.09&98.78$\pm$0.04&97.70$\pm$0.04&99.16$\pm$0.05\\
16   &97.68$\pm$0.05&98.91$\pm$0.04&97.88$\pm$0.05&99.29$\pm$0.04\\
32   &97.80$\pm$0.05&98.97$\pm$0.04&97.94$\pm$0.04&99.29$\pm$0.03\\
64   &97.82$\pm$0.05&98.99$\pm$0.03&97.99$\pm$0.03&{\bf 99.32$\pm$0.02}\\
\hline
\end{tabular}
\par
\end{centering}
\end{small}
\caption{MNIST large network results using data augmentation (Aug) and not using it (No aug).
T stands for the number of ticks used to encode test instances with rate code.}
\label{tab:results-mnist-large}
\end{table}

\section{Discussion}

Results on EEG data show that the performance of the trained network increased from 76\%~\cite{nurse2016acm} to 86\%, which is close to state-of-the-art performance obtained using unconstrained fully connected layers~\cite{nurse2015generalizable}.
Results on MNIST show that the performance of the proposed method is similar compared to previously reported work depending on the configuration.

The use of several ticks for encoding input data shows that a plateau is reached after 8 ticks for MNIST, additional ticks to encode data does not significantly increase performance.

\begin{figure}[htpb]
\centering
  \begin{tabular}{@{}cc@{}}
		\multicolumn{2}{c}{Neuron weight configuration s=[-1,1]} \\
    \includegraphics[width=.45\columnwidth]{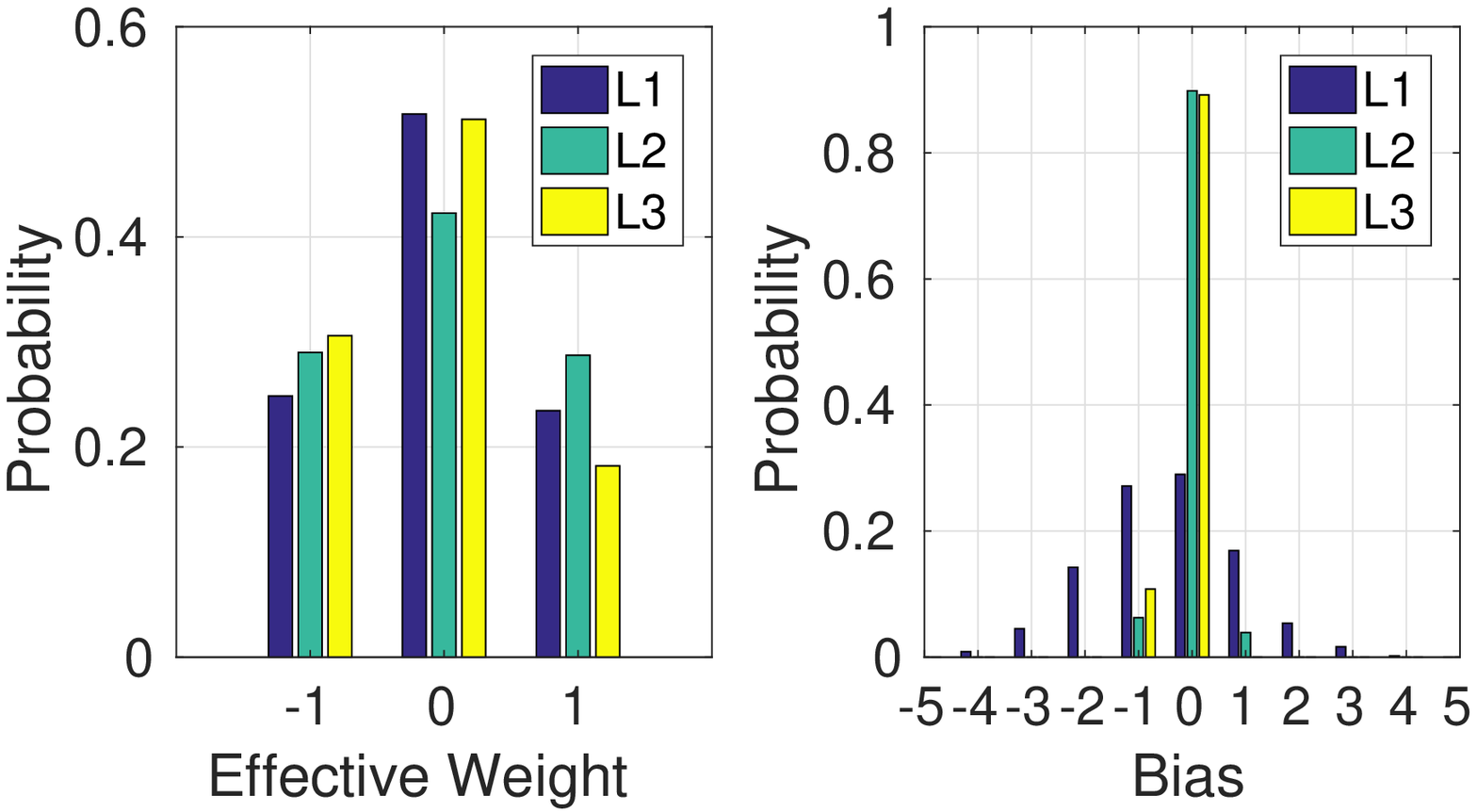} &
    \includegraphics[width=.45\columnwidth]{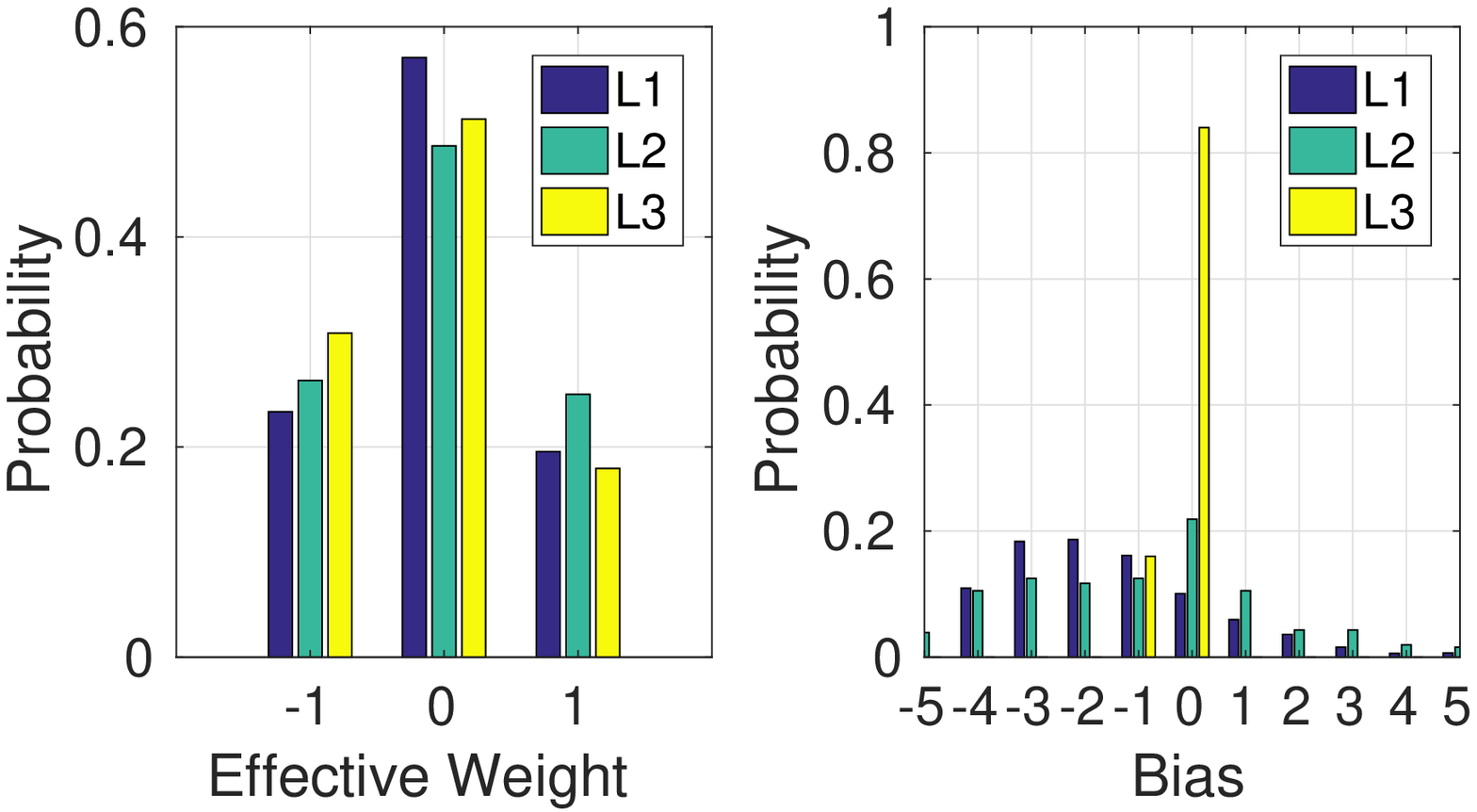} 
    \\
		No augmentation&
		Aug\\
    \multicolumn{2}{c}{Neuron weight configuration s=[-2,-1,1,2]} \\
    \includegraphics[width=.45\columnwidth]{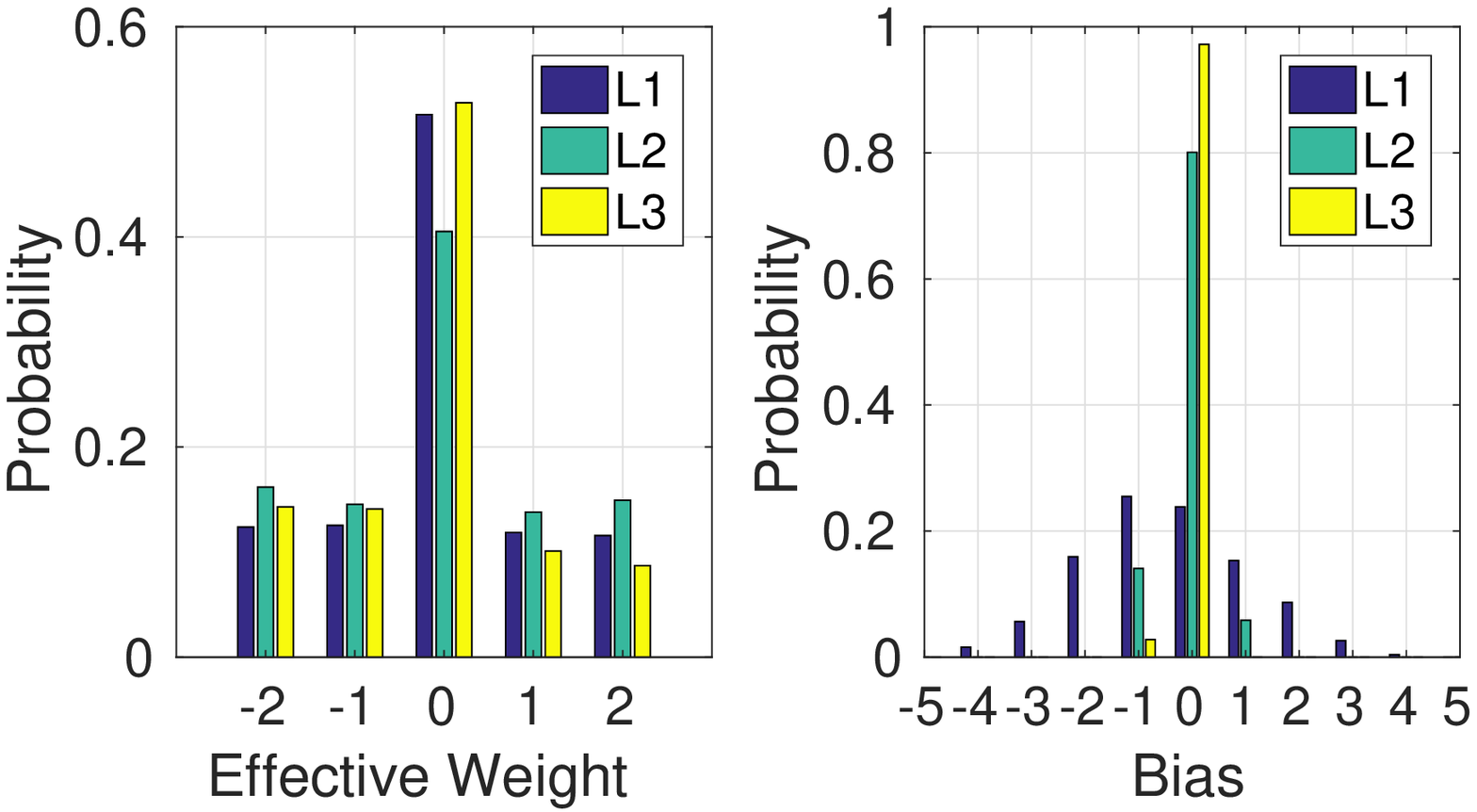} &
    \includegraphics[width=.45\columnwidth]{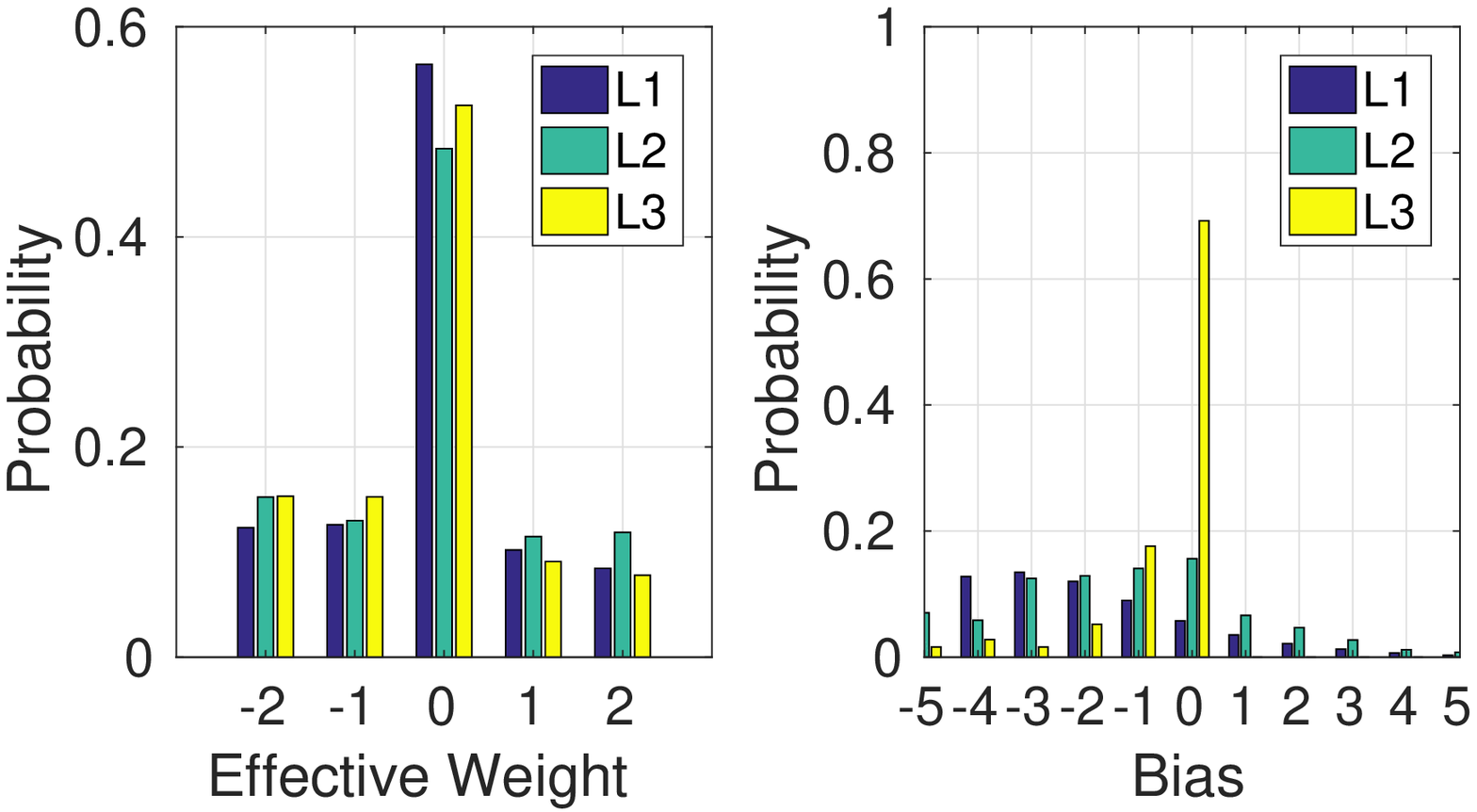} \\
		No augmentation&
		Aug\\
  \end{tabular}
  \caption{Effective weights and bias values for the small network (3 layers, 9 cores) for MNIST data set. Two weight configurations have been used s=[-1,1] and s=[-2,-1,1,2]. The networks have been trained without augmentation (No augmentation) and with augmentation (Aug).}
	\label{fig:small-network-details}
\end{figure}

\begin{figure}[htpb]
\centering
  \begin{tabular}{@{}cc@{}}
		\multicolumn{2}{c}{Neuron weight configuration s=[-1,1]} \\
    \includegraphics[width=.45\columnwidth]{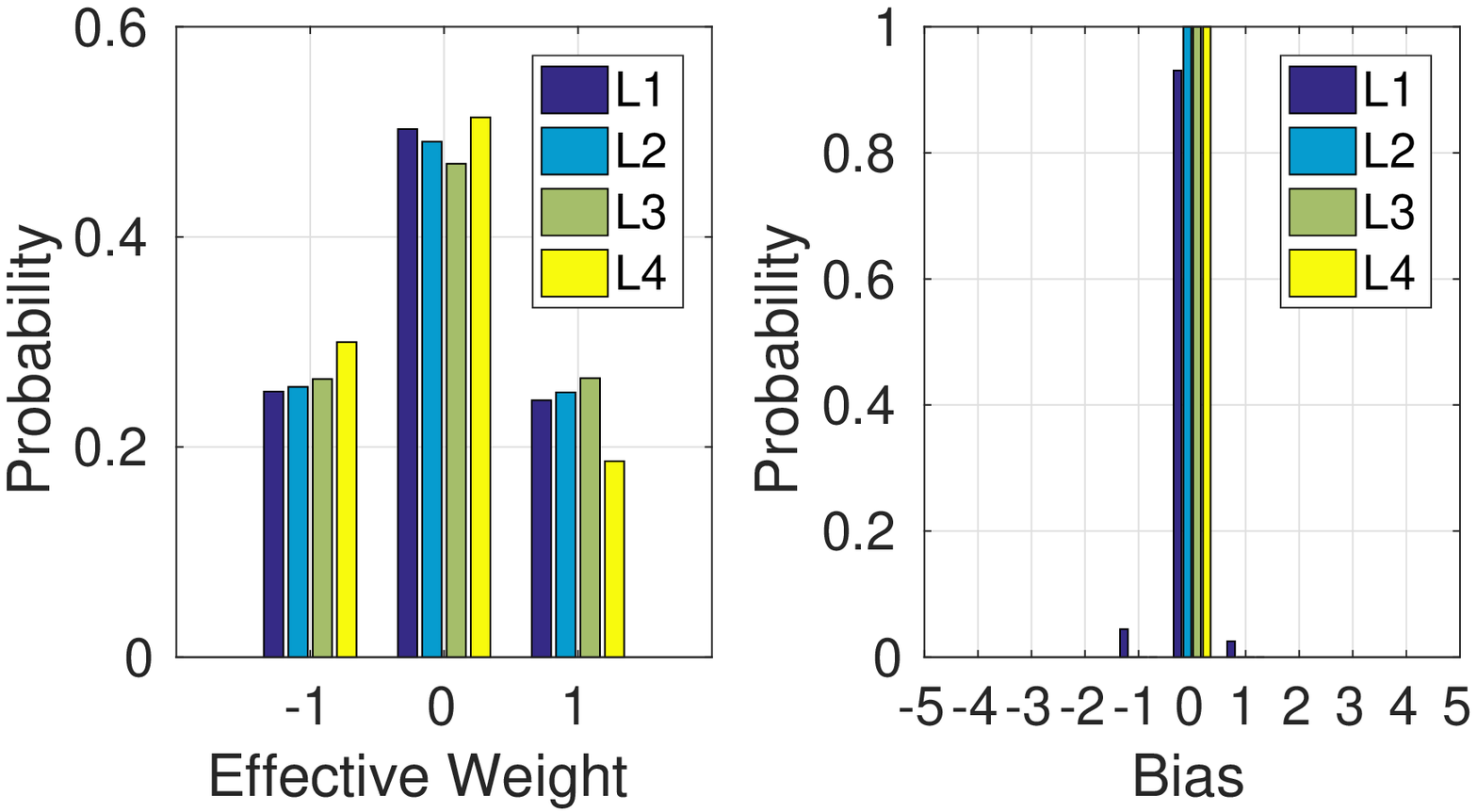} &
    \includegraphics[width=.45\columnwidth]{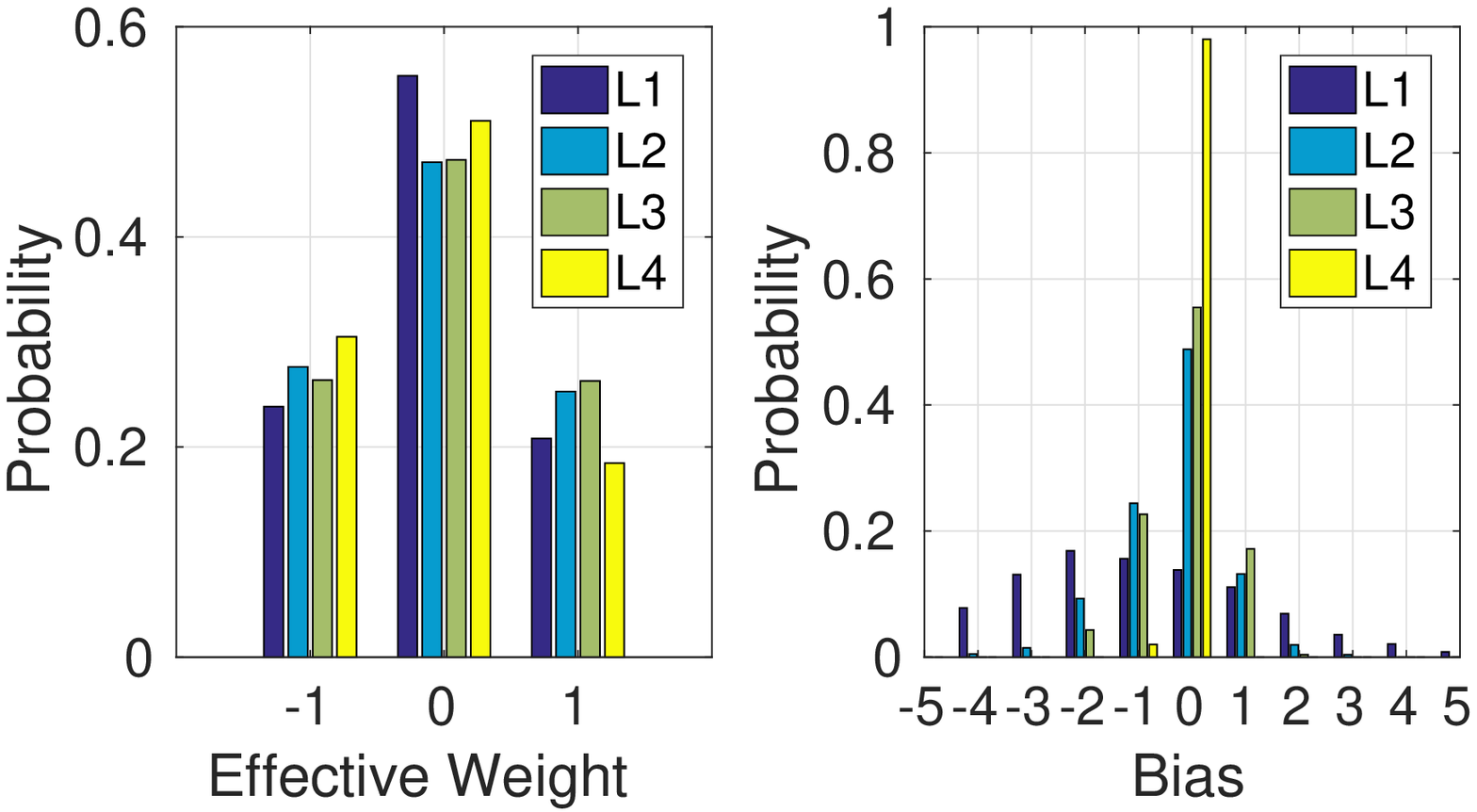} \\
		No augmentation&
		Aug\\
    \multicolumn{2}{c}{Neuron weight configuration s=[-2,-1,1,2]} \\
    \includegraphics[width=.45\columnwidth]{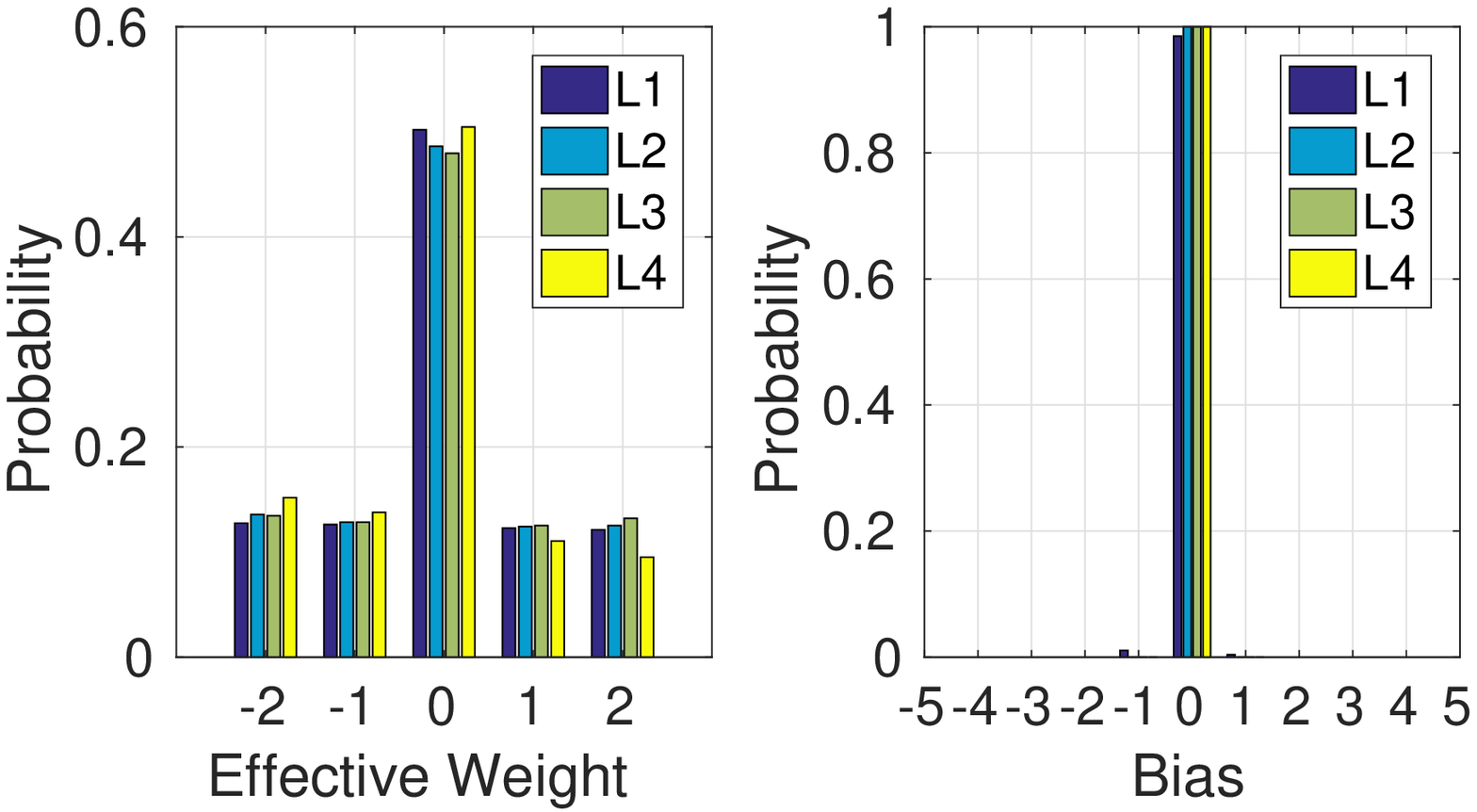} &
    \includegraphics[width=.45\columnwidth]{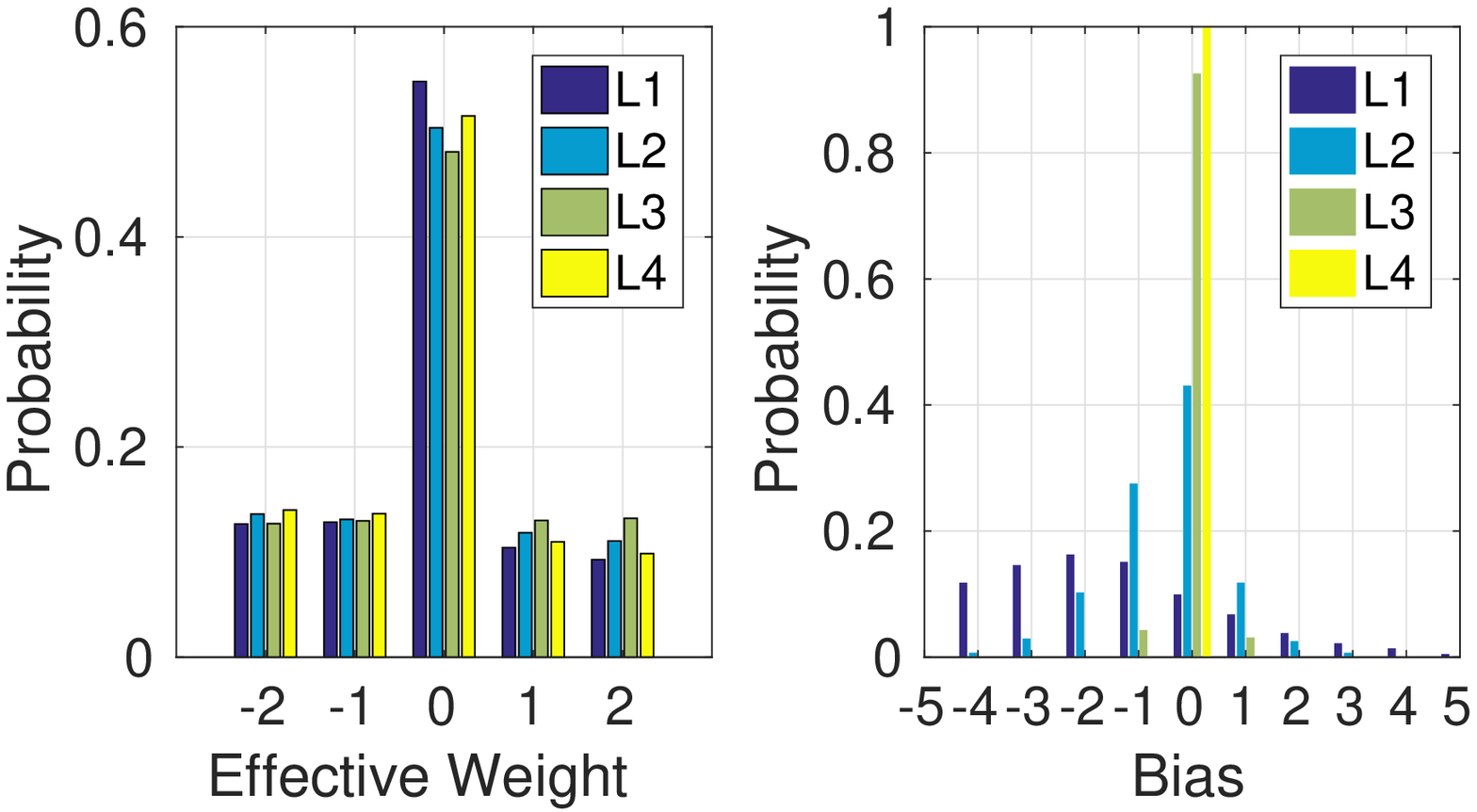} \\
		No augmentation&
		Aug\\
  \end{tabular}
  \caption{Effective weights and bias values for the large network (4 layers, 30 cores) for MNIST data set. Two weight configurations have been used s=[-1,1] and s=[-2,-1,1,2]. The networks have been trained without augmentation (No augmentation) and with augmentation (Aug).}
	\label{fig:large-network-details}
\end{figure}

\begin{figure}[htpb]
\centering
  \begin{tabular}{@{}cc@{}}
		\multicolumn{2}{c}{EEG neuron weight configuration s=[-1,1]} \\
    \includegraphics[width=.45\columnwidth]{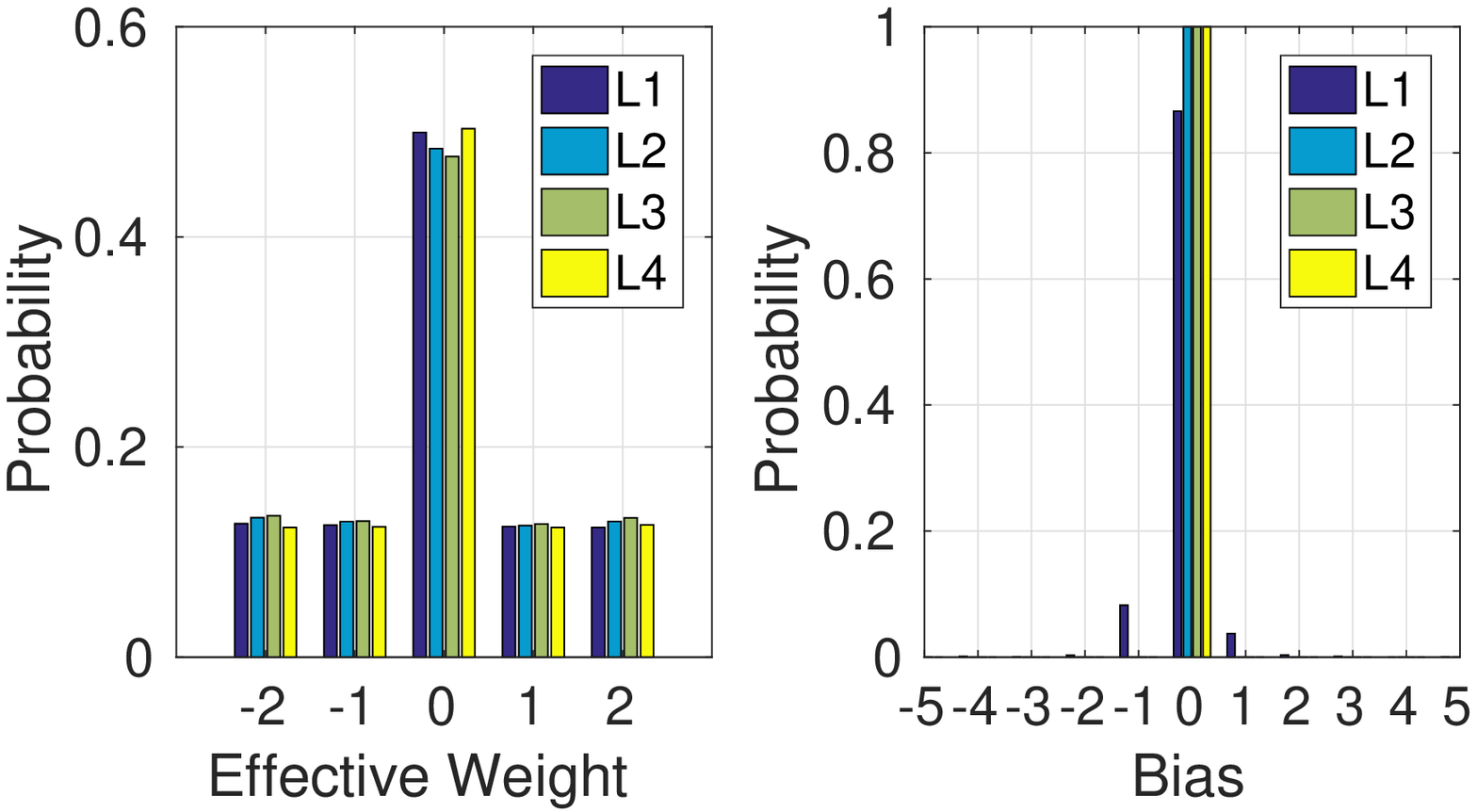} &
    \includegraphics[width=.45\columnwidth]{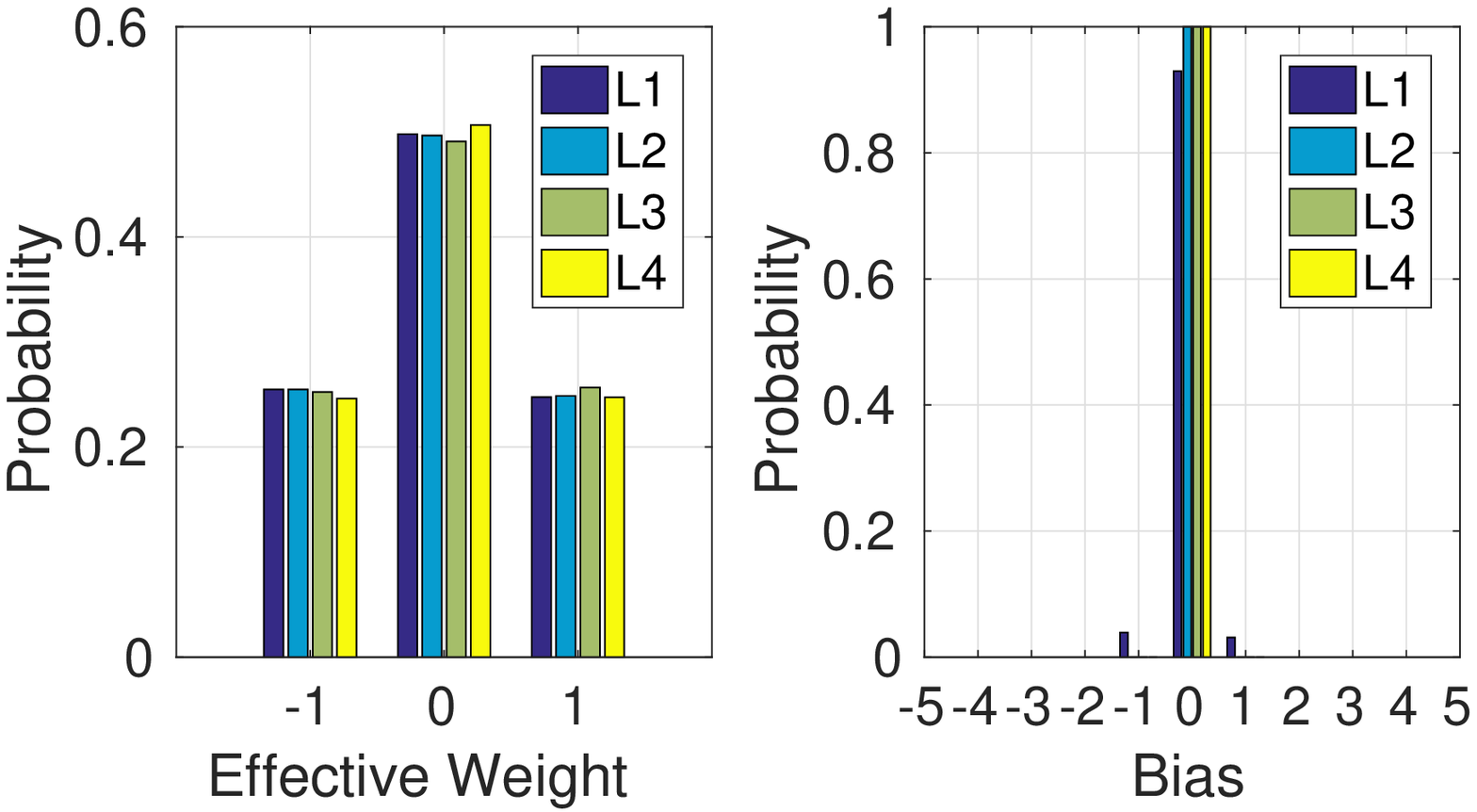} \\
		Original&
		Normalized\\
    \multicolumn{2}{c}{EEG neuron weight configuration s=[-2,-1,1,2]} \\
    \includegraphics[width=.45\columnwidth]{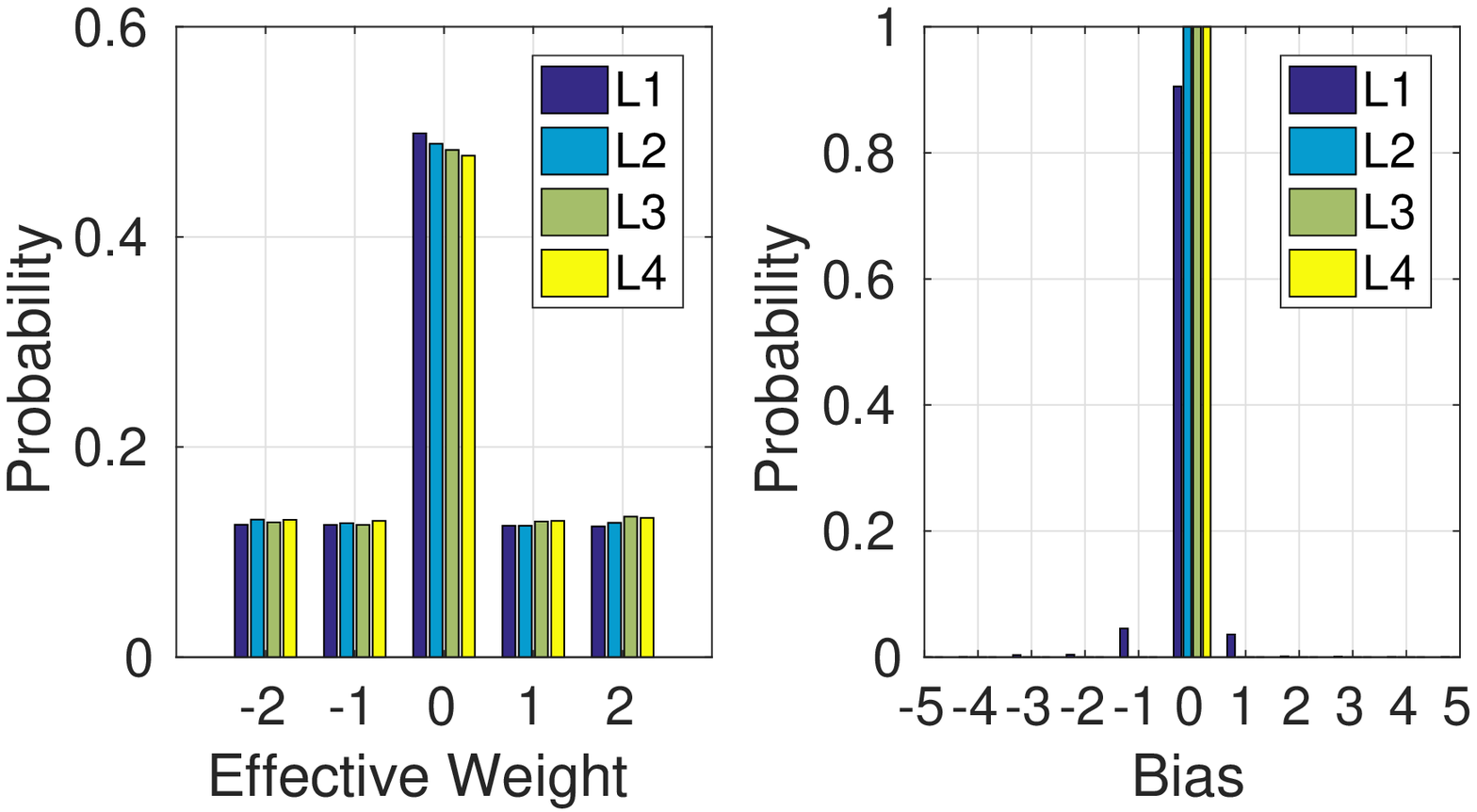} &
    \includegraphics[width=.45\columnwidth]{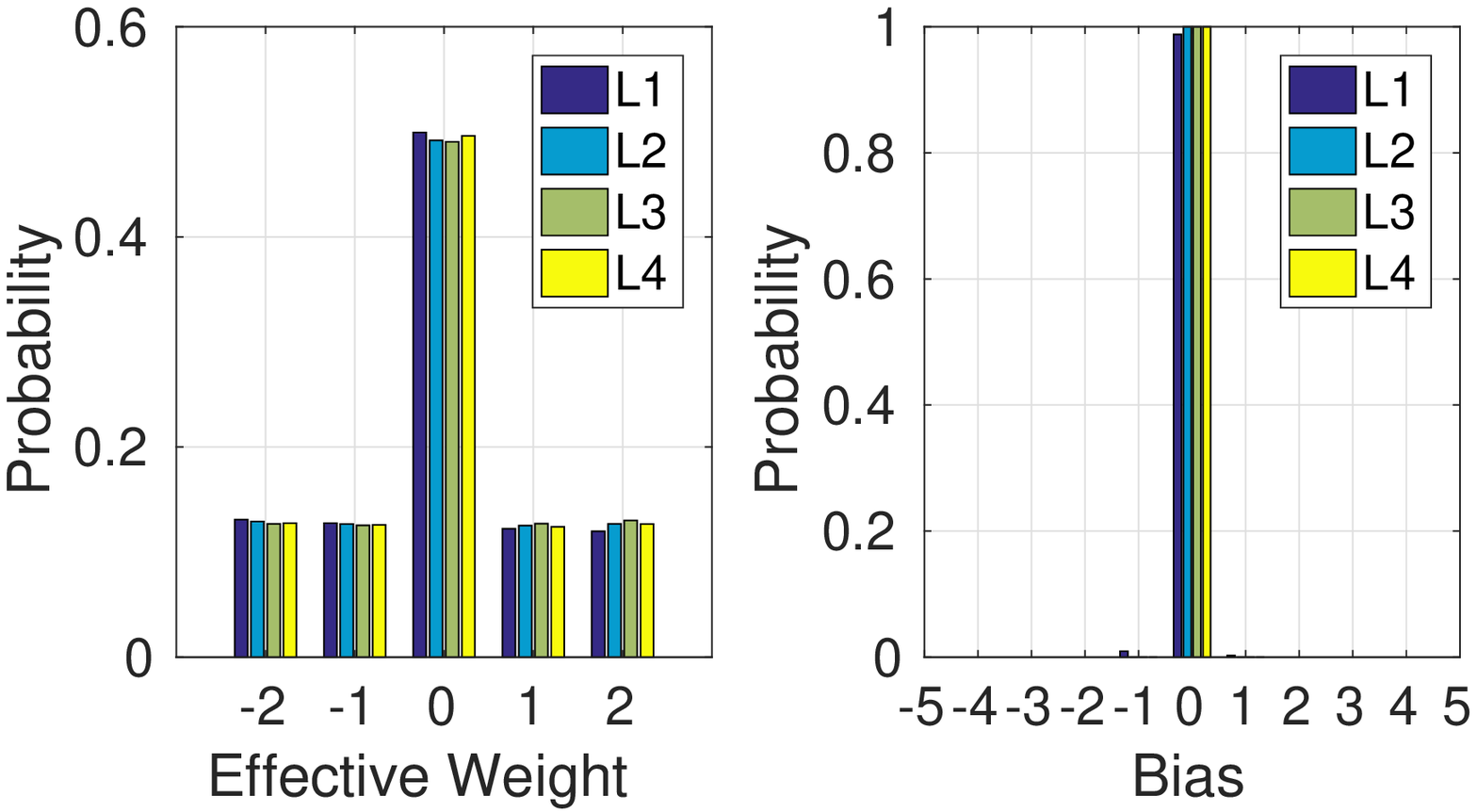} \\
		Original&
		Normalized\\
  \end{tabular}
  \caption{Effective weights and bias values for the large network (4 layers, 30 cores) for EEG data set. Two weight configurations have been used s=[-1,1] and s=[-2,-1,1,2]. The networks have been trained using the original set or the normalized one.}
	\label{fig:eeg-large-network-details}
\end{figure}

We have examined the effective weights and bias values.
Figures~\ref{fig:small-network-details} and~\ref{fig:large-network-details} show the values for MNIST and figure~\ref{fig:eeg-large-network-details} shows the values for the EEG data set.

In all combinations, approximately half of the weights have a value of zero for all the layers, which implies that half of the crossbar connections are disconnected.
The other weights are evenly distributed.
Furthermore, bias values tend to be around zero for all layers in many cases.
On the other hand, when using augmentation in MNIST or in the EEG experiments, bias values tend to be distributed mostly on the negative values for the first layers.
Negative values might indicate that there is a tendency to prevent neurons from spiking.

The values for weights and biases imply that in a deployment scenario, hardware configuration could be further constrained to a limited set of bias values and further research is required to understand implications in hardware design to improve crossbar usage.

\section{Conclusions and future work}

We have shown that it is possible to train constrained fully connected layers on neuromorphic chips for two sets of problems (image classification with MNIST and EEG data analytics), which provides better or similar performance compared to previous work depending on the data set.

Results on EEG data sets which were previously analysed through the original method yield over 86\% classification accuracy using the proposed approach, which is comparable to unconstrained neural networks.
The trained models use a small portion of the TrueNorth chip (30 cores vs. 4096 available in the current version of the chip), thus requiring a much less than 70mW to work, which makes these models suitable for portable autonomous devices with large autonomy.

The outcome of this research provides insights into the possibility of learning network parameters that translate into deployment networks for brain-inspired chips.
We will explore feasibility of expanding the introduced new methodology to train other kinds of existing deep learning algorithms including convolutional neural networks and recurrent neural networks.
Furthermore, analysis of the learnt parameters provide insights that might complement hardware design, thus providing a more efficient deployment of the trained models.

\section{Acknowledgment}
The authors would like to thank the SyNAPSE team at the IBM Almaden Research Center for their support.

\bibliographystyle{named}
\bibliography{bibliography}

\end{document}